\documentclass[runningheads]{llncs}
\usepackage[T1]{fontenc}
\usepackage{graphicx} 
\usepackage{multirow}

\usepackage{amssymb,amsmath}
\usepackage{dsfont}
\usepackage{mathrsfs}
\usepackage{caption}
\usepackage{subcaption}
\usepackage{subfloat}
\usepackage{multirow}
\usepackage{multicol}
\usepackage{diagbox}
\usepackage{comment}
\usepackage{booktabs}
\usepackage{enumitem}
\usepackage{rotating}
\usepackage[hidelinks]{hyperref}

\usepackage{parskip}
\usepackage{natbib}

\newcommand{\E}{\mathbb{E}}
\newcommand{\kl}{D_\text{KL}}

\graphicspath{{figures/}}

\title{Counterfactual Marginalisation: Framework for Evaluating Robustness to Nuisance Variables}
\date{\today}

\begin{document}

\author{Yasin Ibrahim\inst{1} \and Hermione Warr\inst{1} \and \\ Robin J. Evans \inst{2} \and Konstantinos Kamnitsas\inst{1}}

\authorrunning{Y. Ibrahim et al.}
%
\institute{Department of Engineering Science, University of Oxford, Oxford, UK\\
\email{\{firstname.lastname\}@eng.ox.ac.uk} \and Department of Statistics, University of Oxford, Oxford, UK
}
    
\maketitle

\begin{abstract}
Machine learning models can achieve strong test performance while relying on demographic or acquisition-related shortcuts. We propose counterfactual (CF) marginalisation as a test-time evaluation procedure for assessing robustness of classification models to such variables. Given a CF image generator, we intervene on nuisance parent variables such as age or sex, generate CF versions of each test image, and average predictions over a target intervention distribution. This produces intervention-aware predictions that marginalise demographic effects while preserving patient-specific latent information. We use these predictions to define metrics for CF risk, calibration, stability and worst-case sensitivity. We demonstrate this framework's utility for quantitative robustness evaluation.

\end{abstract}

\section{Introduction}
\label{sec:introduction}

Deep learning models have achieved strong performance in medical image analysis, but high accuracy on held-out data does not guarantee clinically appropriate decision rules. In practice, models may rely on demographic, acquisition-related, or site-specific shortcuts rather than pathology-relevant evidence \citep{nofairlunch}. Causal modelling provides a principled way to study these issues by asking, for example, how a prediction would change if a patient's demographic attributes were intervened upon while other patient-specific factors were held fixed \citep{Peters17}. Recent work has used Structural Causal Model (SCM)-based generative models to produce medical image CFs for explanation, auditing and augmentation \citep{deepscm,sanchezdiffusion,hvae,ibrahim2024semi}. Here, instead, we use CFs at test time to define evaluation metrics for medical image classifiers. Given an SCM-based CF generator, we intervene on nuisance parent variables such as age or sex, and generate CF versions of each test image \cite{ibrahim2024semi}. We then develop a range of metrics that use these CFs to measure the sensitivity, stability and performance of a predictor. This setting is closely related to work on fairness and robustness in medical imaging \citep{nofairlunch, kusner2017counterfactual}, where prior studies have shown that models can exhibit demographic disparities and can recover sensitive attributes directly from images \citep{xu2024fairnessreview,readingrace}. Our contributions are as follows:

\begin{itemize}
    \item We formulate \emph{counterfactual marginalisation} over variables as a test-time evaluation procedure for medical image classifiers.
    \item We introduce intervention-aware evaluation metrics, including counterfactual marginal risk, interventional expected risk, counterfactual stability, and worst-case counterfactual risk. This extends counterfactuals from explanation and augmentation to principled robustness evaluation.
    \item We demonstrate that our approach exposes biases in predictors better than traditional metrics and also enables measuring sensitivity to nuisance variables without requiring disease labels. 
\end{itemize}

\section{Background}
\label{sec:background}
Point-estimate evaluation can miss brittle model behaviour: small, plausible changes to inputs or logits may alter predictions. This concern echoes work on flat minima, where models in wider low-loss regions tend to generalise better than sharp solutions \citep{hochreiter1997flat,foret2021sharpness}. Ensembles and Bayesian model averaging address a related issue by marginalising over parameters to improve robustness and uncertainty estimates \citep{kamnitsas2017ensembles,wenzel2020hyperparameter,wilson2020bayesian}, while test-time augmentation similarly averages predictions over transformed images \citep{ayhan2018testtime,kimura2024understanding}. Our counterfactual marginalisation can be viewed as a causal analogue: instead of averaging over ad-hoc image transformations, we average over SCM-defined interventions on parent variables such as age or sex. 
This in turn motivates the design of metrics that assess not only average performance, which often is insufficient for describing clinical utility \cite{maierhein2024metricsreloaded}, but also metrics that measure sensitivity, calibration, and worst-case behaviour across clinically meaningful perturbations.

\section{Counterfactual Marginalisation for Evaluation}
\label{sec:methods}

\begin{figure}[t]
    \centering
    \includegraphics[width=\linewidth]{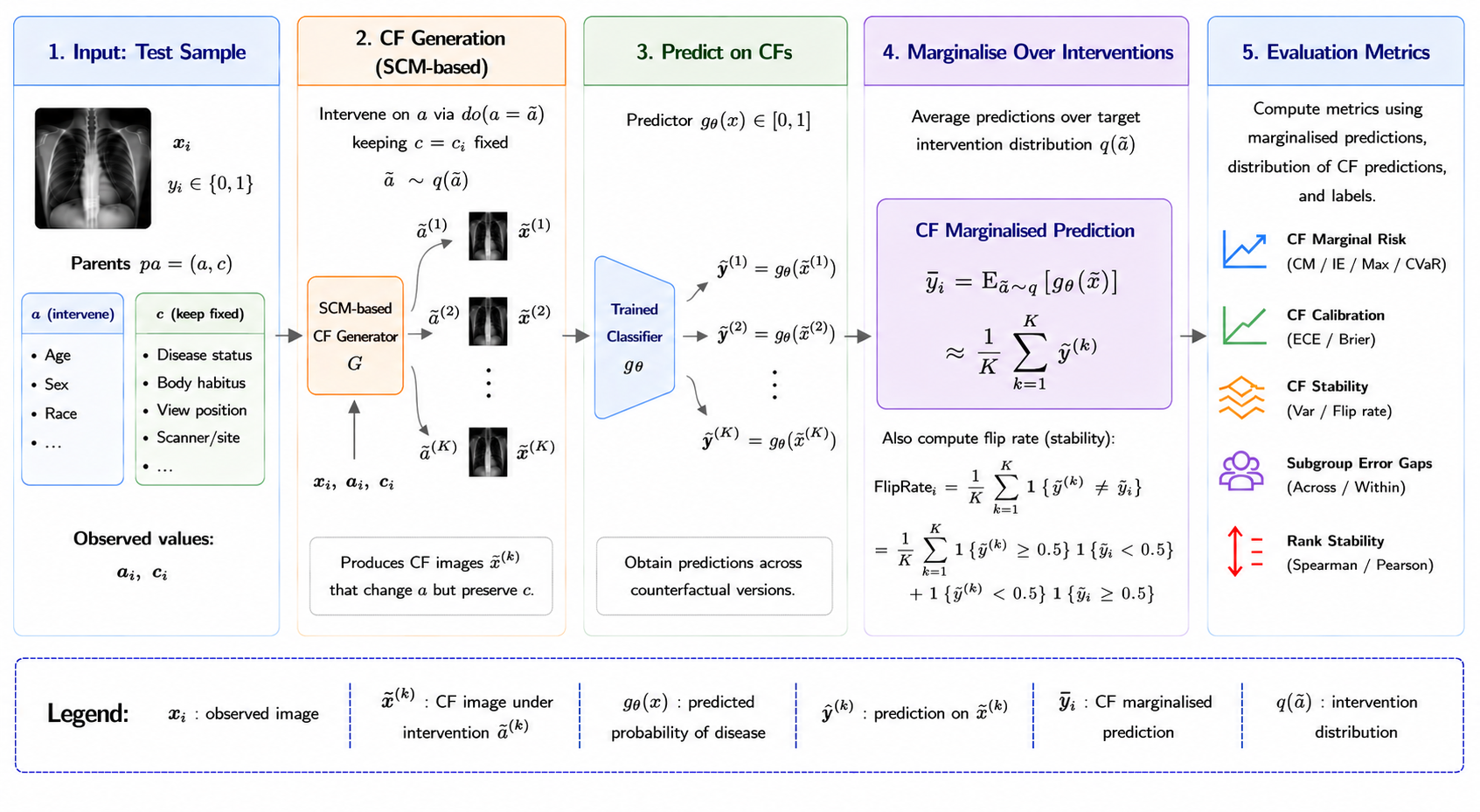}
    \caption{\footnotesize Overview of the proposed Counterfactual Marginalisation evaluation framework} 
    \label{fig:outline}
\end{figure}

Let \(g_\theta(x) \! \in \! [0,1]\) denote the predicted probability of disease for image \(x\) and \(y \in \{0,1\}\) the corresponding ground-truth label. We partition the parent variables as \(\textbf{pa} \! =\! (\mathbf{a}, \mathbf{c})\), where \(\mathbf{a}\) is the subset of parent variables to be marginalised, e.g. age and sex, and \(\mathbf{c}\) denotes the remaining parent variables that are kept fixed. Given an observed sample \((x_i, \mathbf{a}_i, \mathbf{c}_i, y_i)\), we generate counterfactual images by intervening on \(\mathbf{a}\) while preserving the inferred exogenous noise for that individual through the standard abduction--action--prediction procedure. This yields a set of counterfactual images \(
\widetilde{x}_i(\widetilde{\mathbf{a}}) \sim P\big(x \mid \text{do}(\mathbf{a}=\widetilde{\mathbf{a}}), \mathbf{c}_i, u_i \big),\)
where \(u_i\) denotes the abducted exogenous variables for sample \(i\). To remove the influence of the nuisance demographic variables, we define a target distribution \(q(\widetilde{\mathbf{a}})\) over interventions. This may be chosen as the empirical test-set distribution, a uniform distribution over demographic strata, or an external population distribution. We then define the \textbf{counterfactually marginalised prediction} for sample \(i\) as
\begin{equation}
\bar{p}_i = \mathbb{E}_{\widetilde{\mathbf{a}} \sim q} \left[g_\theta\big(\widetilde{x}_i(\widetilde{\mathbf{a}})\big)\right]
\label{eq:marginal_pred}
\end{equation}
In practice, this expectation is approximated by Monte Carlo sampling:
\begin{equation}
\hat{\bar{p}}_i = \frac{1}{M}\sum_{m=1}^{M}g_\theta\big(\widetilde{x}_i(\widetilde{\mathbf{a}}_i^{(m)})\big),
\qquad
\widetilde{\mathbf{a}}_i^{(m)} \sim q(\widetilde{\mathbf{a}})
\label{eq:marginal_pred_mc}
\end{equation}

\subsection{Counterfactual Risk}

Using Eq.~\ref{eq:marginal_pred}, we define the \textbf{counterfactual marginal risk} and \textbf{interventional expected risk} as

\noindent\begin{minipage}{.5\linewidth}
\begin{equation}
\mathcal{R}_{\mathrm{CM}} = \frac{1}{N} \sum_{i=1}^{N} \ell(\bar{p}_i, y_i)
\label{eq:cmr}
\end{equation}
\end{minipage}%
\begin{minipage}{.5\linewidth}
\begin{equation}
\mathcal{R}_{\mathrm{IE}} = \frac{1}{N} \sum_{i=1}^{N} \mathbb{E}_{\widetilde{\mathbf{a}} \sim q} \left[ L_i(\widetilde{\mathbf{a}}) \right]
\label{eq:ier}
\end{equation}
\end{minipage}

where \(L_i(\widetilde{\mathbf{a}}) = \ell\big(g_\theta(\widetilde{x}_i(\widetilde{\mathbf{a}})), y_i\big)\) and \(\ell\) is binary cross-entropy or another appropriate loss. The former evaluates predictive performance after marginalising out the nuisance demographic variables, while the latter averages the loss over counterfactual interventions before aggregation. To measure the extent to which the predictor is vulnerable to the most adverse demographic interventions, we consider a tail-risk measure based on the conditional value at risk (CVaR). For a chosen tail level \(\alpha \in (0,1]\), the \textbf{counterfactual tail risk} is
\begin{equation}
\mathcal{R}_{\mathrm{CVaR},\alpha} = \frac{1}{N} \sum_{i=1}^{N} \mathrm{CVaR}_{\alpha}\big(L_i(\widetilde{\mathbf{a}})\big)
\label{eq:cf_cvar}
\end{equation}
where 
\(\mathrm{CVaR}_{\alpha}\) is the expected loss within the worst \(\alpha\)-fraction of interventions under \(q(\widetilde{\mathbf{a}})\). Let \(\mathcal{R}_{\mathrm{WC}}\! =\! \mathcal{R}_{\mathrm{CVaR},0}\) denote the risk of the worst-case intervention.

\subsection{Counterfactual Calibration}
While \(\mathcal{R}_{\mathrm{CM}}\) and \(\mathcal{R}_{\mathrm{IE}}\) measure predictive accuracy under counterfactual marginalisation, they do not directly assess whether the resulting probabilities are well calibrated. To handle this, we report the \textbf{counterfactual Brier score} and \textbf{counterfactual expected calibration error} (ECE):

\noindent\begin{minipage}{.4\linewidth}
\begin{equation}
  \mathcal{C}_{\mathrm{Brier}} = \frac{1}{N} \sum_{i=1}^{N} (\bar p_i - y_i)^2
\label{eq:cf_brier}
\end{equation}
\end{minipage}%
\begin{minipage}{.6\linewidth}
\begin{equation}
  \mathcal{C}_{\mathrm{ECE}} = \sum_{b=1}^{B} \frac{|I_b|}{N} \left| \frac{1}{|I_b|} \sum_{i \in I_b} y_i - \frac{1}{|I_b|} \sum_{i \in I_b} \bar p_i \right|
\label{eq:cf_ece}
\end{equation}
\end{minipage}

where \(I_1,\dots,I_B\) denote bins of samples partitioned according to their marginalised predicted probabilities \(\bar p_i\). These measure whether the probabilities obtained after marginalisation remain clinically interpretable as calibrated risk estimates.

\subsection{Counterfactual Stability}

 We define a \textbf{counterfactual stability} score based on the variation of the model output across interventions:
\begin{equation}
\mathcal{S}_{\mathrm{var}} = \frac{1}{N} \sum_{i=1}^{N} \mathrm{Var}_{\widetilde{\mathbf{a}} \sim q} \left[ g_\theta\big(\widetilde{x}_i(\widetilde{\mathbf{a}})\big)\right]
\label{eq:stability_var}
\end{equation}
A low value of \(\mathcal{S}_{\mathrm{var}}\) indicates that the predicted probability is stable under changes in the nuisance demographic variables. For threshold-based clinical decision making, we additionally define a \textbf{counterfactual flip rate}
\begin{equation}
\mathcal{S}_{\mathrm{flip}} = \frac{1}{N} \sum_{i=1}^{N} \mathbb{E}_{\widetilde{\mathbf{a}} \sim q} \left[\mathds{1}\left\{\mathds{1}[g_\theta(x_i) > \tau]\neq\mathds{1}[g_\theta(\widetilde{x}_i(\widetilde{\mathbf{a}})) > \tau]\right\}\right]
\label{eq:flip_rate}
\end{equation}
where \(\tau\) is a fixed decision threshold. This metric captures how often a prediction changes solely because of an intervention on the nuisance parent variables. We also measure how often the binary decision made on the observed image differs from the decision made using the counterfactually marginalised prediction. This \textbf{observed--marginalised disagreement rate} is defined as
\begin{equation}
\mathcal{D}_{\mathrm{obs\text{-}marg}} = \frac{1}{N} \sum_{i=1}^{N} \mathds{1} \left\{ \mathds{1}[g_\theta(x_i) > \tau] \neq \mathds{1}[\bar p_i > \tau] \right\}
\label{eq:obs_marg_disagreement}
\end{equation}
This metric quantifies the extent to which conventional observational evaluation would lead to a different clinical decision than intervention-aware evaluation. For tasks in which the predicted score is used for prioritisation or triage, it is important that the relative ranking of patients remains stable after marginalisation \citep{blythe2025continuous, li2021ranking}. Let \(r_i\) denote the rank of the observed prediction \(g_\theta(x_i)\) across the test set, and let \(\bar r_i\) denote the rank of the marginalised prediction \(\bar p_i\). We quantify \textbf{counterfactual rank stability} using the Spearman rank correlation measure:
\begin{equation}
\mathcal{S}_{\mathrm{rank}} = 1-\frac{6\sum_{i=1}^{N}(r_i-\bar r_i)^2}{N(N^2-1)}
\label{eq:cf_rank}
\end{equation}

\section{Theoretical Advantages of Counterfactual Evaluation}
\label{sec:theory}

As well as the clear advantage of being able to generate samples from underrepresented classes, we also have that for quantities that are identified observationally, such as the
mean prediction within a demographic subgroup, the CF estimator achieves
lower variance by exploiting the full test set rather than the subgroup alone. Let $z$ denote a particular attribute value (e.g.\ $A = \text{Female}$),
$n_z = |\{i : a_i = z\}|$ the number of test samples in that subgroup, and
$n$ the total test size.
The standard \emph{observational estimator} of
$\mu_z = \mathbb{E}[g(X) \mid A = z]$ is
\begin{equation}
  \hat{\mu}_z^{\mathrm{obs}} \;=\; \frac{1}{n_z} \sum_{i:\, a_i = z} g(x_i),
  \qquad
  \mathrm{Var}\!\left[\hat{\mu}_z^{\mathrm{obs}}\right]
    = \frac{\mathrm{Var}[g(X) \mid A=z]}{n_z}
  \label{eq:obs_estimator}
\end{equation}
The \emph{CF estimator} generates a counterfactual image at $A=z$ for every
test patient and averages: \(\hat{\mu}_z^{\mathrm{CF}} \;=\; \frac{1}{n} \sum_{i=1}^{n} g\!\left(\tilde{x}_i(z)\right)\). Under correct SCM specification, $\hat{\mu}_z^{\mathrm{CF}}$ is unbiased for $\mu_z$. Its variance decomposes via the law of total variance, conditioning on the abducted exogenous noise $U_i$:
\begin{equation}
  \mathrm{Var}\!\left[\hat{\mu}_z^{\mathrm{CF}}\right]
    = \frac{\mathrm{Var}\!\left[\mathbb{E}[g(\tilde{X}(z)) \mid U]\right]}{n}
    + \frac{\mathbb{E}\!\left[\mathrm{Var}[g(\tilde{X}(z)) \mid U]\right]}{n}
  \label{eq:cf_variance}
\end{equation}
The first term is the variance of the posterior mean (signal); the second
is the expected posterior variance (SCM generation noise $\sigma^2_{\mathrm{gen}}$).
The exogeneous noise \(U\) accounts for all unobserved background variables and individual traits, serving as a sufficient statistic for individual-level causal queries. Then, by the Rao-Blackwell theorem, \(\mathrm{Var}[\mathbb{E}[g(\tilde{X}(z))\! \mid \! U]] \! \leq \!
 \mathrm{Var}[g(\tilde{X}(z))] = \mathrm{Var}[g(X) \! \mid \! A \! = \! z]\), so
\begin{equation}
  \mathrm{Var}\!\left[\hat{\mu}_z^{\mathrm{CF}}\right]
    \;\leq\; \frac{\mathrm{Var}[g(X) \mid A=z]}{n}
             + \frac{\sigma^2_{\mathrm{gen}}}{n}
  \label{eq:cf_variance_bound}
\end{equation}

\begin{proposition}[Variance reduction]
  \label{prop:variance}
  Let $\sigma^2 = \mathrm{Var}[g(X) \mid A=z]$.
  The CF estimator satisfies
  \[
    \mathrm{Var}\!\left[\hat{\mu}_z^{\mathrm{CF}}\right]
    \;\leq\;
    \mathrm{Var}\!\left[\hat{\mu}_z^{\mathrm{obs}}\right]
  \]
  whenever $\sigma^2_{\mathrm{gen}} \leq \sigma^2\!\left(1 - n_z/n\right)$.
  The advantage grows as the subgroup becomes rarer: as $n_z/n \to 0$,
  the bound reduces to $\mathrm{Var}[\hat{\mu}_z^{\mathrm{CF}}] \leq
  \sigma^2/n + \sigma^2_{\mathrm{gen}}/n \ll \sigma^2/n_z$.
\end{proposition}

This is directly analogous to the efficiency gain of a within-subjects design over a between-subjects design in experimental settings \citep{sibbald1998crossover}.

\section{Experiments}
\label{sec:experiments}

\subsection{Datasets}
\label{sec:experiments:datasets}
To demonstrate the utility of our metrics, we use two large scale chest x-ray datasets, CheXpert \citep{chexpert} and MIMIC-CXR \citep{mimic}. We perform binary classification of \emph{Pleural Effusion} (PE), existence or not. To ensure a controlled evaluation, we balance the test split so that male and female patients appear in equal numbers, and disease prevalence is equalised across sex and 3 age groups (<45, 45-65, >65). 
The final balanced test sets for CheXpert (C) and MIMIC (M) contain $5{,}180$ and $28{,}189$ studies respectively. We use these two datasets to run three experimental settings: (1) Train and test the PE classifier on C; (2) Train and test on M; (3) Train on C and test on M$_{nat}$ with natural demographic distributions. For all 3 settings we train the CF generator on C.

\subsection{Predictors and Training Settings}
\label{sec:experiments:model}

The causal graph encodes the following parents of the image: sex, age, race, lung size, and pleural effusion. All experiments vary the \emph{sex} attribute ($\tilde{a} \in
\{\text{Male}, \text{Female}\}$) and \emph{age} (8
uniformly-spaced values from 20 to 90 years), yielding $K=2 \times 8 = 16$ counterfactual images per patient. The marginalisation distribution $q(\tilde{A})$ is uniform over the $K$ intervention values in all experiments.
The baseline predictor (\texttt{basic}) is trained on the full training set without any demographic manipulation. To study how well our method assesses bias, we create a first set of biased predictors by adding a fixed scalar offset $\beta \in \{0.5, 1.0, 2.0\}$
to the baseline logit for female patients at test time: \(g_\beta(x, s) \!=\! \sigma\!\left(\text{logit}(g_0(x)) + \beta \cdot s\right)\), where $s \in \{0,1\}$ is the sex label.
Because the offset is a monotone transformation applied uniformly within each sex group, within-group rankings are preserved and hence stratified AUC is invariant to $\beta$ by construction.
We then train three sex-biased predictors on a corrupted dataset in which Male$+$disease and Female$+$healthy samples are dropped with probabilities $0.8$, $0.9$, and $0.95$ respectively, inducing a spurious correlation between sex and PE. We also train predictors with a continuous age bias that induces a correlation between older age and positive PE.
For each training sample with normalised age $a \in [0,1]$ and binary PE label $y$, the concordance score is defined as
\(
  c_i \!=\! a_i \cdot \mathbf{1}[y_i = 1]
           + (1 - a_i) \cdot \mathbf{1}[y_i = 0],
\)
and a sample is removed from training with probability
$\beta_{\mathrm{age}} \cdot (1 - c_i)$,
where $\beta_{\mathrm{age}} \in \{0.8, 0.9, 0.95\}$ controls the strength
of the induced correlation. 


\subsection{Results}
\label{sec:experiments:results}

\begin{table}[h]
\centering
\caption{\footnotesize Marginal risk decomposition -- PE classifier is trained and tested on CheXpert.}
\begin{tabular}{l|ccccccc}
Predictor & {$R_\text{orig}$} & {$R_\text{CM}$} & {$R_\text{IE}$} & {$R_\text{CVaR}(0.50)$} & {$R_\text{CVaR}(0.25)$} & {$R_\text{CVaR}(0.10)$} & {$R_\text{WC}$}\\
\hline
Basic & 0.444 & 0.501 & 0.594 & 1.002 & 1.326 & 1.549 & 1.852\\
\hline
Logit $+0.5$ & 0.445 & 0.612 & 0.724 & 1.027 & 1.346 & 1.597 & 1.845 \\
Logit $+1$ & 0.443 & 0.629 & 0.771 & 1.068 & 1.424 & 1.699 & 1.962 \\
Logit $+2$ & 0.502 & 0.677 & 0.919 & 1.072 & 1.431 & 1.716 & 1.997 \\
\hline
Sex ($\alpha=0.80$) & 0.453 & 0.610 & 0.687 & 1.192 & 1.618 & 1.991 & 1.390 \\
Sex ($\alpha=0.90$) & 0.452 & 0.614 & 0.699 & 1.195 & 1.612 & 1.968 & 2.338 \\
Sex ($\alpha=0.95$) & 0.443 & 0.614 & 0.699 & 1.203 & 1.622 & 1.982 & 2.361 \\
\hline
Age ($\alpha=0.80$) & 0.494 & 0.613 & 0.750 & 1.126 & 1.511 & 1.817 & 2.111 \\
Age ($\alpha=0.90$) & 0.509 & 0.612 & 0.754 & 1.225 & 1.654 & 2.007 & 2.340 \\
Age ($\alpha=0.95$) & 0.512 & 0.613 & 0.758 & 1.518 & 2.046 & 2.495 & 2.893 \\
\hline
CTC ($\uparrow$) & 6.33 & 8.33 & \textbf{8.66} & 8.33 & 8.00 & 8.00 & 7.33\\
\end{tabular}
\label{tab:chex:risk}
\end{table}

Table \ref{tab:chex:risk} shows that in the CheXpert experiment (Setting 1), our CF metrics discern the biases in the various predictors better than the standard risk $R_\text{orig}$ (cross entropy) calculated only on original data, which sometimes decreases when bias increases (e.g. Logit$+0.5\rightarrow$ Logit$+1$). To summarise comparisons, we define the Correct Transition Count (CTC) as the number of transitions with increasing predictor bias (out of 9 total: 3 for logit bias, $Basic \! \rightarrow \! Logit\! + \! 0.5 \rightarrow \! + \! 1 \! \rightarrow \! + \! 2$, and similarly 3 for sex and age bias) where the risk also increases. Higher is better and counterfactual metrics outperform the standard risk estimation $R_{orig}$. All results shown are averages of 3 repetitions per experiment with varying seeds. Similar results were found for setting 2 and 3 (See Appendix \ref{app:results} for details).

\begin{figure}[h]
    \centering
    \includegraphics[width=\linewidth]{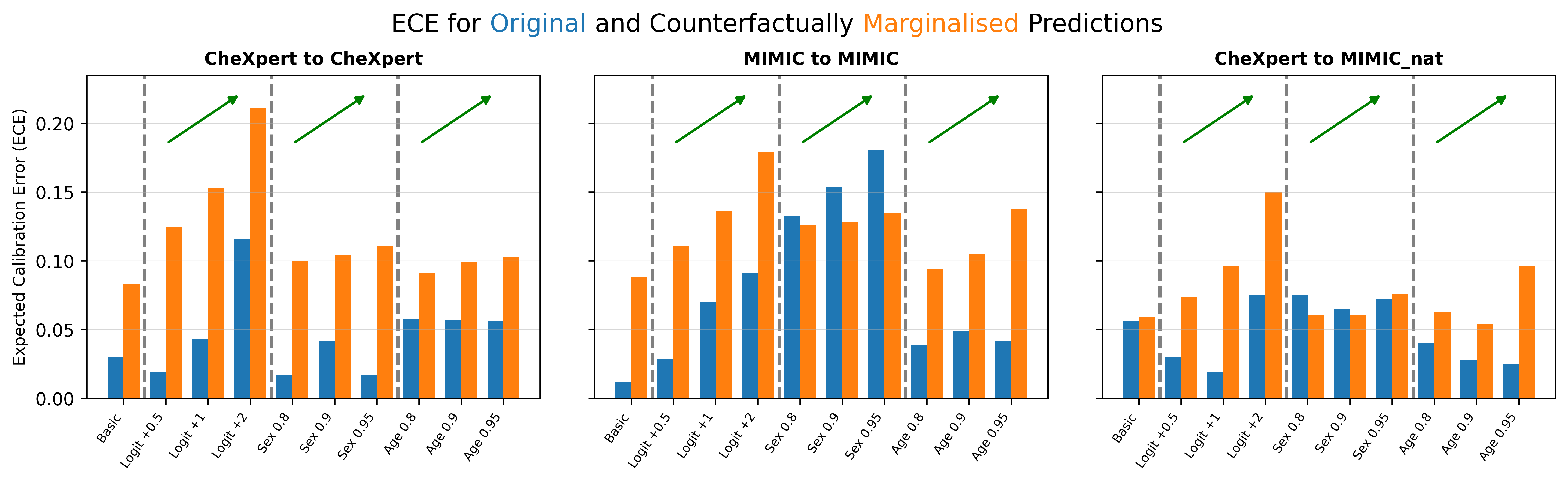}
    \caption{\footnotesize Model calibration: Marginalised ECE increases more consistently with predictor bias than the original ECE, suggesting it is a better measure of calibration.}
    \label{fig:calibration}
\end{figure}

Moreover, biased predictors should be worse calibrated \citep{li2021overfitting}. Fig. \ref{fig:calibration} compares the original ECE \cite{guo2017calibration} and marginalised ECE (Eq.\ref{eq:cf_ece}) for each predictor, where we find that scores of the latter increase more consistently with increasing predictor bias. This suggests that CF marginalised ECE can better estimate model calibration. 

\begin{figure}[h]
    \centering
    \includegraphics[width=\linewidth]{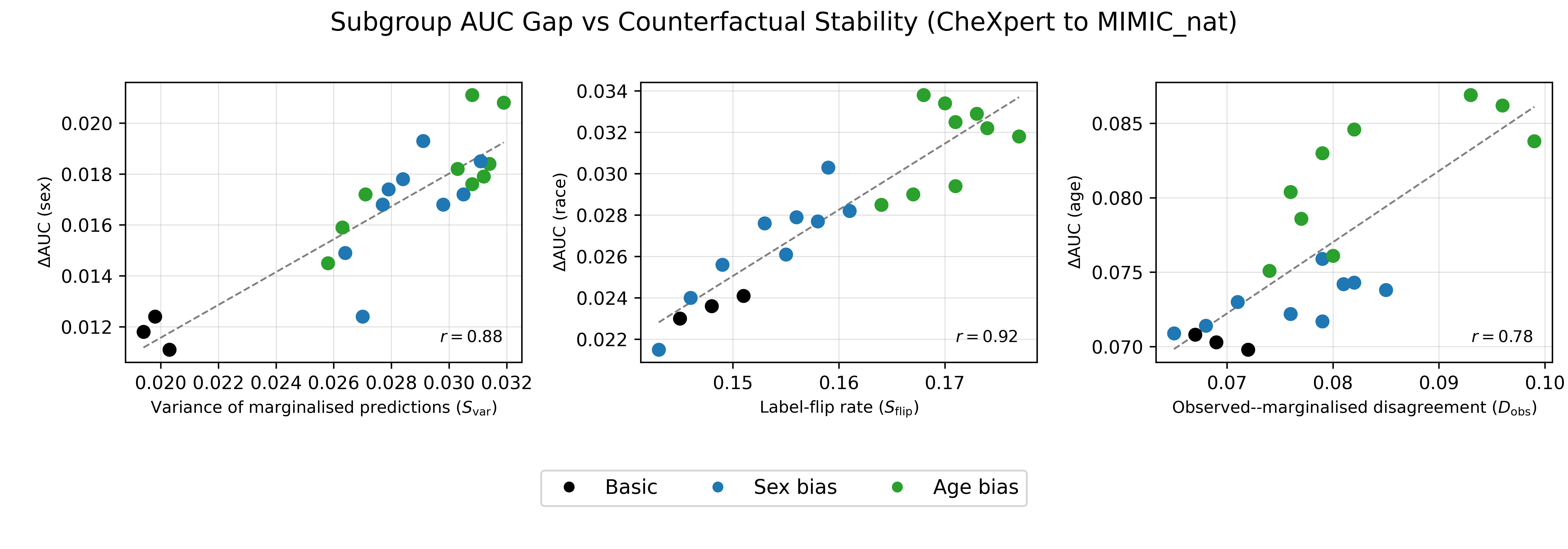}
    \caption{\footnotesize Our stability metrics correlate strongly with AUC gaps, suggesting they are a useful proxy when labels are unavailable at test time.}
    \label{fig:stability}
\end{figure}

Finally, we assess our stability metrics. 
When ground truth disease labels are available, bias is often estimated by calculating the performance gap between the best and worst performing data stratification for a variable of interest. Therefore, separately for each attribute (sex, race, age), we calculate the worst-case gap in AUC ($\Delta AUC$) between relevant data stratifications (e.g. male/female for sex) for the 1 non-biased (basic) predictor and each of the 3 sex-biased and 3 age-biased predictors. We repeat for 3 random seeds and thus obtain 21 values of $\Delta AUC$ per attribute (sex, race, age). We then study how well our stability metrics correlate with $\Delta AUC$ and Fig. \ref{fig:stability} demonstrates strong correlation, implying our metrics are useful bias estimators.
\
Table \ref{tab:c2m:stability} further shows that values of our stability metrics tend to follow changes in model bias.
\
These results suggests that \emph{our metrics can be used as bias estimators without requiring ground truth classification labels}, replacing the use of stratified performance gap analysis, which requires ground truth labels, when these are not available.

\begin{table}[h]
\centering
\caption{\footnotesize Stability under marginalisation for cross-dataset experiment. As desired, increase in bias leads in most cases to increase of $S_{var}$, $S_{flip}$, $D_{obs}$, and decrease of $S_{rank}$. This consistent relation indicate that the metrics can be used to estimate levels of bias.}
\begin{tabular}{l|cccc}
Predictor & {$S_\text{var}$} & {$S_\text{flip}$} & {$D_\text{obs}$} & $\mathcal{S}_{\mathrm{rank}}$\\
\hline
Basic & 0.0198 & 0.143 & 0.069 & 0.920 \\
\hline
Logit $+0.5$ & 0.0212 & 0.144 & 0.064 & 0.918 \\
Logit $+1$ & 0.0359 & 0.188 & 0.115 & 0.907 \\
Logit $+2$ & 0.0502 & 0.243 & 0.173 & 0.847 \\
\hline
Sex ($\alpha=0.80$) & 0.0270 & 0.146 & 0.068 & 0.916 \\
Sex ($\alpha=0.90$) & 0.0305 & 0.156 & 0.079 & 0.906 \\
Sex ($\alpha=0.95$) & 0.0284 & 0.158 & 0.082 & 0.903 \\
\hline
Age ($\alpha=0.80$) & 0.0263 & 0.167 & 0.077 & 0.917 \\
Age ($\alpha=0.90$) & 0.0308 & 0.170 & 0.079 & 0.916 \\
Age ($\alpha=0.95$) & 0.0314 & 0.174 & 0.096 & 0.901 \\
\end{tabular}
\label{tab:c2m:stability}
\end{table}

\section{Conclusion}
\label{sec:conclusion}

This work introduced \emph{counterfactual marginalisation} as a test-time framework for evaluating medical image classifiers under controlled interventions on nuisance parent variables. By averaging predictions over SCM-generated counterfactual variants, the framework provides intervention-aware measures of risk, calibration, stability, tail sensitivity, and ranking robustness. Results on CheXpert and MIMIC demonstrate that counterfactual evaluation can expose sensitivity to demographic interventions even when standard point-estimate metrics appear stable. A limitation is that the metrics depend on the realism and causal validity of the counterfactual generator, so they should be used alongside generator validation and conventional external evaluation, rather than as a replacement. Future work should study sensitivity to causal graph specification, intervention distributions, generator uncertainty, and extensions to multi-label settings.

\bibliographystyle{plain}
\bibliography{bib}

\clearpage

\appendix
\section{Full Results}
\label{app:results}
\subsection{Risk}

\begin{table}[ht]
\centering
\caption{\footnotesize Marginal risk decomposition (uniform prior). $R_\text{orig}$ = risk of original predictor; $R_\text{CM}$ = risk of the complete-marginalisation predictor; $R_\text{IE}$ = individual-environment risk; $R_\text{CVaR}(\alpha)$ = conditional value-at-risk at level $\alpha$; $R_\text{WC}$ = worst-case risk; CTC ($\uparrow$) = correct transition count.}
\label{tab:risk}

\begin{minipage}{\textwidth}
\centering
\subcaption{\footnotesize Trained and tested on CheXpert ($N = 5180$).}
\label{tab:chex:risk}
\scalebox{0.68}{
\begin{tabular}{l|ccccccc}
Predictor & {$R_\text{orig}$} & {$R_\text{CM}$} & {$R_\text{IE}$} & {$R_\text{CVaR}(0.50)$} & {$R_\text{CVaR}(0.25)$} & {$R_\text{CVaR}(0.10)$} & {$R_\text{WC}$}\\
\hline
Basic & 0.444 & 0.501 & 0.594 & 1.002 & 1.326 & 1.549 & 1.852\\
\hline
Logit $+0.5$ & 0.445 & 0.612 & 0.724 & 1.027 & 1.346 & 1.597 & 1.845 \\
Logit $+1$ & 0.443 & 0.629 & 0.771 & 1.068 & 1.424 & 1.699 & 1.962 \\
Logit $+2$ & 0.502 & 0.677 & 0.919 & 1.072 & 1.431 & 1.716 & 1.997 \\
\hline
Sex ($\alpha=0.80$) & 0.453 & 0.610 & 0.687 & 1.192 & 1.618 & 1.991 & 1.390 \\
Sex ($\alpha=0.90$) & 0.452 & 0.614 & 0.699 & 1.195 & 1.612 & 1.968 & 2.338 \\
Sex ($\alpha=0.95$) & 0.443 & 0.614 & 0.699 & 1.203 & 1.622 & 1.982 & 2.361 \\
\hline
Age ($\alpha=0.80$) & 0.494 & 0.613 & 0.750 & 1.126 & 1.511 & 1.817 & 2.111 \\
Age ($\alpha=0.90$) & 0.509 & 0.612 & 0.754 & 1.225 & 1.654 & 2.007 & 2.340 \\
Age ($\alpha=0.95$) & 0.512 & 0.613 & 0.758 & 1.518 & 2.046 & 2.495 & 2.893 \\
\hline
CTC ($\uparrow$) & 6.33 & 8.33 & \textbf{8.66} & 8.33 & 8.00 & 8.00 & 7.33\\
\end{tabular}}
\end{minipage}

\vspace{0.1em}

\begin{minipage}{\textwidth}
\centering
\subcaption{\footnotesize Trained and tested on MIMIC ($N = 28189$).}
\label{tab:mimic:risk}
\scalebox{0.68}{
\begin{tabular}{l|ccccccc}
Predictor & {$R_\text{orig}$} & {$R_\text{CM}$} & {$R_\text{IE}$} & {$R_\text{CVaR}(0.50)$} & {$R_\text{CVaR}(0.25)$} & {$R_\text{CVaR}(0.10)$} & {$R_\text{WC}$}\\
\hline
Basic & 0.335 & 0.425 & 0.536 & 0.790 & 1.031 & 1.190 & 1.283\\
\hline
Logit $+0.5$ & 0.342 & 0.454 & 0.556 & 0.827 & 1.084 & 1.252 & 1.351 \\
Logit $+1$ & 0.359 & 0.469 & 0.590 & 0.897 & 1.161 & 1.370 & 1.494 \\
Logit $+2$ & 0.424 & 0.511 & 0.694 & 0.880 & 1.181 & 1.377 & 1.505 \\
\hline
Sex ($\alpha=0.80$) & 0.375 & 0.455 & 0.570 & 1.055 & 1.290 & 1.459 & 1.563 \\
Sex ($\alpha=0.90$) & 0.371 & 0.458 & 0.583 & 1.215 & 1.427 & 1.580 & 1.681 \\
Sex ($\alpha=0.95$) & 0.384 & 0.459 & 0.581 & 1.232 & 1.499 & 1.724 & 1.864 \\
\hline
Age ($\alpha=0.80$) & 0.335 & 0.425 & 0.536 & 0.857 & 1.160 & 1.360 & 1.549 \\
Age ($\alpha=0.90$) & 0.351 & 0.565 & 0.734 & 0.887 & 1.142 & 1.370 & 1.501 \\
Age ($\alpha=0.95$) & 0.366 & 0.610 & 0.771 & 0.880 & 1.191 & 1.411 & 1.505 \\
\hline
CTC ($\uparrow$) & 7.00 & \textbf{8.33} & 8.00 & 8.00 & 7.66 & 7.00 & 7.33 \\
\end{tabular}}
\end{minipage}

\vspace{0.1em}

\begin{minipage}{\textwidth}
\centering
\subcaption{\footnotesize Trained on CheXpert, tested on MIMIC ($N = 28189$).}
\label{tab:c2m:risk}
\scalebox{0.68}{
\begin{tabular}{l|ccccccc}
Predictor & {$R_\text{orig}$} & {$R_\text{CM}$} & {$R_\text{IE}$} & {$R_\text{CVaR}(0.50)$} & {$R_\text{CVaR}(0.25)$} & {$R_\text{CVaR}(0.10)$} & {$R_\text{WC}$}\\
\hline
Basic & 0.397 & 0.466 & 0.549 & 0.769 & 0.977 & 1.121 & 1.210 \\
\hline
Logit $+0.5$ & 0.402 & 0.486 & 0.566 & 0.822 & 1.055 & 1.223 & 1.328 \\
Logit $+1$ & 0.422 & 0.495 & 0.602 & 0.913 & 1.176 & 1.379 & 1.506 \\
Logit $+2$ & 0.508 & 0.536 & 0.722 & 1.177 & 1.506 & 1.778 & 1.945 \\
\hline
Sex ($\alpha=0.80$) & 0.418 & 0.497 & 0.558 & 0.761 & 0.941 & 1.064 & 1.140 \\
Sex ($\alpha=0.90$) & 0.395 & 0.499 & 0.565 & 0.793 & 1.023 & 1.183 & 1.281 \\
Sex ($\alpha=0.95$) & 0.437 & 0.530 & 0.597 & 0.829 & 1.042 & 1.193 & 1.287 \\
\hline
Age ($\alpha=0.80$) & 0.416 & 0.490 & 0.568 & 0.897 & 1.056 & 1.143 & 1.464 \\
Age ($\alpha=0.90$) & 0.422 & 0.511 & 0.573 & 0.809 & 1.015 & 1.166 & 1.563 \\
Age ($\alpha=0.95$) & 0.402 & 0.516 & 0.614 & 0.827 & 1.062 & 1.233 & 1.642 \\
\hline
CTC ($\uparrow$) & 6.66 & 8.00 & \textbf{8.33} & 8.33 & 7.66 & 7.66 & 7.33 \\
\end{tabular}}
\end{minipage}
\end{table}

\clearpage

\subsection{Calibration}
\begin{table}[ht]
\centering
\caption{Calibration before and after marginalisation (10 bins). ``Marginalised'' = marginalised predictions; ``Original'' = original predictions. Brier = Brier score; ECE = expected calibration error; MCE = maximum calibration error.}
\label{tab:calibration}

\begin{minipage}[t]{0.49\textwidth}
\centering
\subcaption{Trained and tested on CheXpert ($N = 5180$).}
\label{tab:chex:calibration}
\resizebox{\linewidth}{!}{%
\begin{tabular}{lrrr|rrr}

& \multicolumn{3}{c|}{Marginalised} & \multicolumn{3}{c}{Original} \\
\cmidrule(lr){2-4}\cmidrule(l){5-7}
Predictor & {Brier} & {ECE} & {MCE} & {Brier} & {ECE} & {MCE} \\
\hline
Basic & 0.202 & 0.083 & 0.171 & 0.143 & 0.030 & 0.080 \\
\hline
Logit $+0.5$ & 0.208 & 0.125 & 0.219 & 0.143 & 0.019 & 0.077 \\
Logit $+1$ & 0.216 & 0.153 & 0.240 & 0.149 & 0.043 & 0.121 \\
Logit $+2$ & 0.237 & 0.211 & 0.306 & 0.180 & 0.116 & 0.228 \\
\hline
Sex ($\alpha=0.80$) & 0.218 & 0.100 & 0.184 & 0.146 & 0.017 & 0.046 \\
Sex ($\alpha=0.90$) & 0.223 & 0.104 & 0.232 & 0.145 & 0.042 & 0.119 \\
Sex ($\alpha=0.95$) & 0.224 & 0.111 & 0.233 & 0.142 & 0.017 & 0.078 \\
\hline
Age ($\alpha=0.80$) & 0.216 & 0.091 & 0.164 & 0.158 & 0.058 & 0.104 \\
Age ($\alpha=0.90$) & 0.216 & 0.099 & 0.159 & 0.162 & 0.057 & 0.089 \\
Age ($\alpha=0.95$) & 0.228 & 0.103 & 0.171 & 0.165 & 0.056 & 0.079 \\

\end{tabular}}
\end{minipage}
\hfill
\begin{minipage}[t]{0.49\textwidth}
\centering
\subcaption{Trained and tested on MIMIC ($N = 28189$).}
\label{tab:mimic:calibration}
\resizebox{\linewidth}{!}{%
\begin{tabular}{lrrr|rrr}

& \multicolumn{3}{c|}{Marginalised} & \multicolumn{3}{c}{Original} \\
\cmidrule(lr){2-4}\cmidrule(l){5-7}
Predictor & {Brier} & {ECE} & {MCE} & {Brier} & {ECE} & {MCE} \\
\hline
Basic & 0.139 & 0.088 & 0.244 & 0.102 & 0.012 & 0.067 \\
\hline
Logit $+0.5$ & 0.143 & 0.111 & 0.284 & 0.104 & 0.029 & 0.093 \\
Logit $+1$ & 0.150 & 0.136 & 0.323 & 0.109 & 0.050 & 0.146 \\
Logit $+2$ & 0.167 & 0.179 & 0.378 & 0.128 & 0.091 & 0.212 \\
\hline
Sex ($\alpha=0.80$) & 0.152 & 0.126 & 0.299 & 0.193 & 0.133 & 0.262 \\
Sex ($\alpha=0.90$) & 0.186 & 0.128 & 0.303 & 0.212 & 0.154 & 0.271 \\
Sex ($\alpha=0.95$) & 0.207 & 0.135 & 0.296 & 0.241 & 0.181 & 0.328 \\
\hline
Age ($\alpha=0.80$) & 0.141 & 0.094 & 0.251 & 0.113 & 0.039 & 0.149 \\
Age ($\alpha=0.90$) & 0.150 & 0.105 & 0.252 & 0.114 & 0.049 & 0.188 \\
Age ($\alpha=0.95$) & 0.154 & 0.138 & 0.259 & 0.118 & 0.042 & 0.073 \\

\end{tabular}}
\end{minipage}

\vspace{1em}

\begin{minipage}[t]{0.49\textwidth}
\centering
\subcaption{Trained on CheXpert and tested on MIMIC ($N = 28189$).}
\label{tab:c2m:calibration}
\resizebox{\linewidth}{!}{%
\begin{tabular}{lrrr|rrr}

& \multicolumn{3}{c|}{Marginalised} & \multicolumn{3}{c}{Original} \\
\cmidrule(lr){2-4}\cmidrule(l){5-7}
Predictor & {Brier} & {ECE} & {MCE} & {Brier} & {ECE} & {MCE} \\
\hline
Basic & 0.155 & 0.059 & 0.103 & 0.125 & 0.056 & 0.147 \\
\hline
Logit $+0.5$ & 0.157 & 0.074 & 0.161 & 0.126 & 0.030 & 0.077 \\
Logit $+1$ & 0.161 & 0.096 & 0.206 & 0.134 & 0.019 & 0.034 \\
Logit $+2$ & 0.179 & 0.150 & 0.278 & 0.164 & 0.075 & 0.146 \\
\hline
Sex ($\alpha=0.80$) & 0.156 & 0.061 & 0.116 & 0.132 & 0.075 & 0.163 \\
Sex ($\alpha=0.90$) & 0.169 & 0.061 & 0.139 & 0.125 & 0.065 & 0.182 \\
Sex ($\alpha=0.95$) & 0.172 & 0.076 & 0.132 & 0.140 & 0.072 & 0.187 \\
\hline
Age ($\alpha=0.80$) & 0.159 & 0.063 & 0.141 & 0.131 & 0.040 & 0.075 \\
Age ($\alpha=0.90$) & 0.167 & 0.064 & 0.137 & 0.135 & 0.028 & 0.083 \\
Age ($\alpha=0.95$) & 0.170 & 0.096 & 0.163 & 0.127 & 0.015 & 0.048 \\

\end{tabular}}
\end{minipage}
\end{table}

\clearpage
\subsection{Stability}

\begin{table}[ht]
\centering
\caption{Prediction stability under marginalisation at threshold $\tau=0.5$ (uniform prior). $S_\text{var}$ = variance of marginalised predictions; $S_\text{flip}$ = label-flip rate at $\tau$; $D_\text{obs}$ = mean absolute difference between original and marginalised scores; $\rho$ = Spearman rank correlation; $\tau_b$ = Kendall $\tau_b$; $|\Delta\bar{r}|$ = mean absolute rank change.}
\label{tab:stability}

\begin{minipage}[t]{0.49\textwidth}
\centering
\subcaption{Trained and tested on CheXpert ($N = 5180$).}
\label{tab:chex:stability}
\resizebox{\linewidth}{!}{%
\begin{tabular}{l|cccccc}
Predictor & {$S_\text{var}$} & {$S_\text{flip}$} & {$D_\text{obs}$} & {$\rho$} & {$\tau_b$} & {$|\Delta\bar{r}|$}\\
\hline
Basic & 0.0396 & 0.247 & 0.199 & 0.691 & 0.476 &  913.5 \\
\hline
Logit $+0.5$ & 0.0530 & 0.271 & 0.216 & 0.653 & 0.471 &  918.4\\
Logit $+1$ & 0.0574 & 0.289 & 0.235 & 0.653 & 0.469 &  928.5\\
Logit $+2$ & 0.0685 & 0.335 & 0.291 & 0.641 & 0.466 &  934.2\\
\hline
Sex ($\alpha=0.80$) & 0.0426 & 0.265 & 0.218 & 0.653 & 0.476 &  920.6\\
Sex ($\alpha=0.90$) & 0.0501 & 0.256 & 0.208 & 0.645 & 0.479 &  922.4\\
Sex ($\alpha=0.95$) & 0.0489 & 0.263 & 0.222 & 0.644 & 0.464 &  932.5 \\
\hline
Age ($\alpha=0.80$) & 0.0566 & 0.267 & 0.203 & 0.652 & 0.477 &  913.3\\
Age ($\alpha=0.90$) & 0.0567 & 0.270 & 0.205 & 0.642 & 0.466 &  937.8\\
Age ($\alpha=0.95$) & 0.0567 & 0.279 & 0.208 & 0.589 & 0.416 & 1050.2\\
\end{tabular}}
\end{minipage}
\hfill
\begin{minipage}[t]{0.49\textwidth}
\centering
\subcaption{Trained and tested on MIMIC ($N = 28189$).}
\label{tab:mimic:stability}
\resizebox{\linewidth}{!}{%
\begin{tabular}{l|cccccc}
Predictor & {$S_\text{var}$} & {$S_\text{flip}$} & {$D_\text{obs}$} & {$\rho$} & {$\tau_b$} & {$|\Delta\bar{r}|$}\\
\hline
Basic & 0.0358 & 0.148 & 0.069 & 0.932 & 0.783 & 2143 \\
\hline
Logit $+0.5$ & 0.0367 & 0.148 & 0.071 & 0.930 & 0.782 & 2059\\
Logit $+1$ & 0.0385 & 0.153 & 0.080 & 0.909 & 0.760 & 2242\\
Logit $+2$ & 0.0433 & 0.170 & 0.108 & 0.877 & 0.745 & 2731\\
\hline
Sex ($\alpha=0.80$) & 0.0406 & 0.158 & 0.102 & 0.919 & 0.758 & 2041\\
Sex ($\alpha=0.90$) & 0.0497 & 0.196 & 0.137 & 0.897 & 0.713 & 3013\\
Sex ($\alpha=0.95$) & 0.0514 & 0.191 & 0.139 & 0.865 & 0.671 & 3386 \\
\hline
Age ($\alpha=0.80$) & 0.0412 & 0.155 & 0.087 & 0.928 & 0.773 & 2244\\
Age ($\alpha=0.90$) & 0.0477 & 0.151 & 0.089 & 0.917 & 0.770 & 2285\\
Age ($\alpha=0.95$) & 0.0487 & 0.168 & 0.097 & 0.888 & 0.746 & 2330\\
\end{tabular}}
\end{minipage}

\vspace{1em}

\begin{minipage}[t]{0.49\textwidth}
\centering
\subcaption{Trained on CheXpert and tested on MIMIC ($N = 28189$).}
\label{tab:c2m:stability}
\resizebox{\linewidth}{!}{%
\begin{tabular}{l|cccccc}
Predictor & {$S_\text{var}$} & {$S_\text{flip}$} & {$D_\text{obs}$} & {$\rho$} & {$\tau_b$} & {$|\Delta\bar{r}|$}\\
\hline
Basic & 0.0358 & 0.148 & 0.069 & 0.920 & 0.765 & 2339 \\
\hline
Logit $+0.5$ & 0.0212 & 0.144 & 0.064 & 0.918 & 0.751 & 2489 \\
Logit $+1$ & 0.0359 & 0.188 & 0.115 & 0.907 & 0.726 & 2790 \\
Logit $+2$ & 0.0502 & 0.243 & 0.173 & 0.847 & 0.646 & 3801 \\
\hline
Sex ($\alpha=0.80$) & 0.0270 & 0.146 & 0.068 & 0.916 & 0.756 & 2430 \\
Sex ($\alpha=0.90$) & 0.0305 & 0.156 & 0.079 & 0.906 & 0.743 & 2552\\
Sex ($\alpha=0.95$) & 0.0284 & 0.158 & 0.082 & 0.903 & 0.738 & 2605 \\
\hline
Age ($\alpha=0.80$) & 0.0263 & 0.167 & 0.077 & 0.917 & 0.750 & 2380 \\
Age ($\alpha=0.90$) & 0.0308 & 0.170 & 0.079 & 0.916 & 0.748 & 2403 \\
Age ($\alpha=0.95$) & 0.0314 & 0.174 & 0.096 & 0.901 & 0.735 & 2431 \\
\end{tabular}}
\end{minipage}
\end{table}

\clearpage
\section{Details of the Generative Model}
\label{app:model}

For completeness, we summarise the semi-supervised deep structural causal model of
\cite{ibrahim2024semi}, which we use throughout this work to generate counterfactual
images. We refer the reader to the original paper for full derivations; here we
restate only what is needed to make our experiments reproducible.

\begin{figure}[h]
    \centering
    \includegraphics[width=\textwidth]{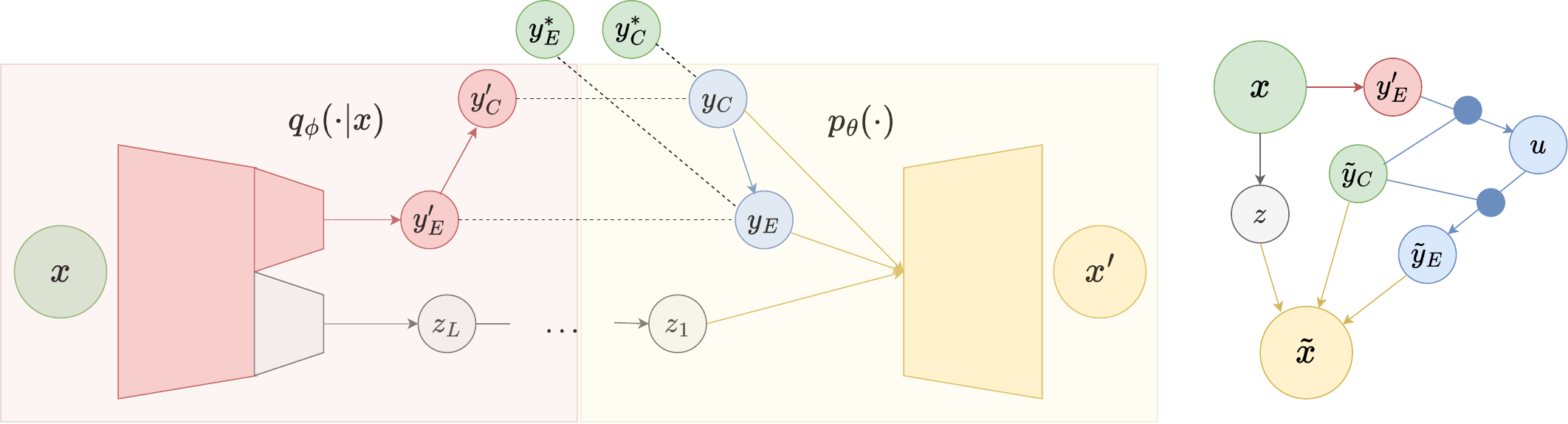}
    \caption{\footnotesize Outline of the model of \cite{ibrahim2024semi}, reproduced here for
    reference. Green: observed, grey: latent, red: predicted, blue: causal generative,
    yellow: decoding. (left) Training; the \(y\) predictions are used for decoding unless
    the labels are observed. (right) Counterfactual generation.}
    \label{fig:app:model}
\end{figure}

\subsection{Setup}
The endogenous variables comprise an image \(x\) and a set of causal variables \(y\),
denoted \(y^*\) when observed and \(y'\) when predicted. Latent variables
\(z = z_{1:L}\) form part of the exogenous noise for \(x\) and are modelled with a
hierarchical latent structure \citep{hvae, ladder}. In addition to this generative
backbone, the model contains a predictive component that infers \(y\) from \(x\),
which is what permits counterfactual generation when labels are missing. As in
\cite{ibrahim2024semi}, we present the case of a single cause variable \(y_C\) and a
single effect variable \(y_E\); the construction extends to any finite number of
variables under an arbitrary causal structure.

\subsection{Training Objective}
Training proceeds by partitioning the data according to which labels are available.
For \textbf{fully labelled} samples drawn from \(\mathcal{D}_L\), the ELBO gives

\begin{align*}
    \log p_\theta(x) \; &\geq \; \E_{q_\phi(z|x,y_E,y_C)}
    \left[\log \frac{p_\theta(x|z,y_E,y_C)\,p_\theta(z)\,p_\theta(y_E|y_C)\,p_\theta(y_C)}
    {q_\phi(z|x,y_E,y_C)}\right] \\
    \Rightarrow \; \mathcal{S}(x,y) \; &:= \; -\mathcal{L}(x,y) - \log p_\theta(y_E|y_C)
    - \log p_\theta(y_C),
\end{align*}
where \(\mathcal{L}(x,y)\) is the ELBO of a conditional VAE and \(p_\theta(\cdot)\)
are Gaussian priors. For \textbf{unlabelled} samples from \(\mathcal{D}_U\), the labels
are predicted via \(y' \sim q_\phi(y|x)\) and regularised against their priors:
\begin{align}
\begin{split}
    \mathcal{U}(x) := -\E_{q_\phi(y_E|x)}\big[&\E_{q_\phi(y_C|x,y_E)}(\mathcal{L}(x,y)
    - \kl\{q_\phi(y_C|x,y_E)||p_\theta(y_C)\})\big]\\
    + &\kl\{q_\phi(y_E|x)||p_\theta(y_E|y_C)\}.
\end{split}
\label{eq:app:bothunlab}
\end{align}
When \textbf{only the cause \(\mathbf{y_C}\)} is observed, for \((x,y_C) \in \mathcal{D}_C\),
\begin{equation}
\mathcal{C}(x,y_C) := -\E_{q_\phi(y_E|x)}[\mathcal{L}(x,y)] - \log p_\theta(y_C)
+ \kl\{q_\phi(y_E|x)||p_\theta(y_E|y_C)\},
\label{eq:app:cause}
\end{equation}
and when \textbf{only the effect \(\mathbf{y_E}\)} is observed, for
\((x,y_E) \in \mathcal{D}_E\),
\begin{equation*}
\mathcal{E}(x,y_E) := -\E_{q_\phi(y_C|x,y_E)}[\mathcal{L}(x,y) + \log p_\theta(y_E|y_C)]
+ \kl\{q_\phi(y_C|x,y_E)||p_\theta(y_C)\}.
\end{equation*}
Since the parent predictors \(q_\phi(y_E|x)\) and \(q_\phi(y_C|x,y_E)\) would otherwise
only receive gradient when their targets are unobserved, an explicit classification term
is added on the labelled subset \citep{semisup}, yielding the total objective

\begin{align}
\begin{split}
\mathcal{T}(x,y) := &\sum_{(x,y)\in \mathcal{D}_L} \mathcal{S}(x,y)
+ \sum_{x \in \mathcal{D}_U}\mathcal{U}(x)
+ \hspace{-1em}\sum_{(x,y_C)\in \mathcal{D}_C}\hspace{-0.5em} \mathcal{C}(x,y_C)\\
+ &\sum_{(x,y_E)\in \mathcal{D}_E} \hspace{-0.5em}\mathcal{E}(x,y_E)
- \E_{(x,y) \in \mathcal{D}_L} \left[\log q_\phi(y_i|y_{<i},x)\right].
\end{split}
\label{eq:app:total}
\end{align}

In the final term the labelled variables \(y_i\) are placed in a topological ordering
\citep{topolog} beginning at the root nodes, so that for every \(i\) the ancestors of
\(y_i\) lie in \(y_{<i}\) and its descendants in \(y_{>i}\).

Two key implementation details of \cite{ibrahim2024semi} are retained. First, for discrete variables whose true label \(y^*\) is unavailable, the corresponding term is inversely
weighted by the entropy \(H_\phi(y'|x)\) of the predicted label distribution --- e.g.\
the expectation in \eqref{eq:app:cause} is scaled by \(1 - H_\phi(y'_E|x)\) --- using
predictive uncertainty as a proxy for how far the imputed label should be trusted.
Second, the expectations over labels in \eqref{eq:app:bothunlab} are computed by
summing over all values of a discrete \(y\),
\(\E_{q_\phi(y|x)}[f(y,\cdot)] = \sum_y q_\phi(y|x)\, f(y,\cdot)\) \citep{semisup},
which becomes expensive as the number of variables or classes grows. Rather than
resorting to Monte-Carlo estimation, training begins on the labelled subset alone, so
that the predictors \(q_\phi(\cdot|x)\) are already accurate before being used to
impute labels for the remainder of the data.

\subsection{Counterfactual Regularisation}
The model additionally employs a causal form of consistency regularisation
\citep{consistency1}, in which perturbations of the input are restricted to
interventions on the DAG over the causal variables. Given an image \(x\) with variables
\((y_C, y_E)\), intervening on the effect via \(\text{do}(y_E = \tilde{e})\) produces
\(\tilde{x}\) with variables \((\tilde{y}_C, \tilde{y}_E)\). If the DAG holds, the cause
is invariant to this intervention, so predicting \(\tilde{y}_C \sim q_\phi(\cdot|\tilde{x})\)
and minimising \(D(y_C, \tilde{y}_C)\) --- for a suitable distance \(D(\cdot,\cdot)\) ---
penalises violations of that invariance; if \(y_C\) is unobserved it is first imputed as
\(y_C \sim q_\phi(\cdot|x,y_E)\). Intervening on the cause instead should change the
effect: \(\tilde{y}_E \sim q_\phi(\cdot|\tilde{x})\) is compared against the counterfactual
value of \(y_E\) under the intervention on \(y_C\), with \(y_E \sim q_\phi(\cdot|x)\)
imputed when unlabelled. This regularises both the generative component and the causal
mechanisms \(p_\theta(y_i|y_{<i})\).

\subsection{Counterfactual Generation}
At inference time, counterfactuals \(\tilde{x}\) are produced using the causal
mechanisms \(p_\theta(y_i|y_{<i})\). Abduction requires the structural assignments to be
invertible in the exogenous noise, so each mechanism is encoded as an invertible map
\(g_{y_{<i}}(u)\), \(u \sim p(u)\), parameterised by the causal parents \(y_{<i}\)
\citep{deepscm}. Each counterfactual value is then obtained in topological order as
\(\tilde{y}_i = g_{\tilde{y}_{<i}}\big(g_{y_{<i}}^{-1}(y_i)\big)\), and any missing
\(y_i\) is imputed beforehand with the trained predictor \(q_\phi(y_i|x,y_{>i})\), as
illustrated in Fig.~\ref{fig:app:model} (right).

\clearpage 

\section{Example Counterfactuals}
Below we provide some example counterfactuals produced by the model and used for testing with our proposed metrics.

\begin{figure}
    \begin{subfigure}[]{\textwidth}
        \includegraphics[width=\textwidth]{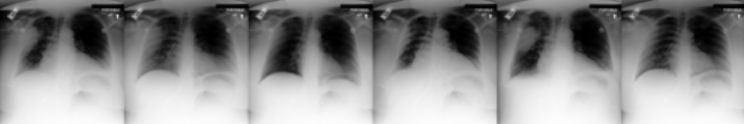}
        \caption{From left to right: (1) original: white, diseased, 52-year-old male (2) do(age=22) (3) do(healthy) (4) do(black) (5) do(female) (6) do(all).}
    \end{subfigure}
    \vskip\baselineskip
    \begin{subfigure}[]{\textwidth}
        \includegraphics[width=\textwidth]{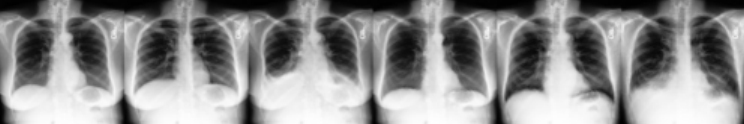}
        \caption{(1) original: black, healthy, 77-year-old female (2) do(age=20) (3) do(diseased) (4) do(white) (5) do(male) (6) do(all).}
    \end{subfigure}
    \vskip\baselineskip
    \begin{subfigure}[]{\textwidth}
        \includegraphics[width=\textwidth]{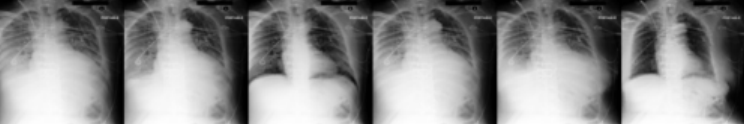}
        \caption{(1) original: white, diseased, 52-year-old male (2) do(age=82) (3) do(healthy) (4) do(asian) (5) do(female) (6) do(all).}
    \end{subfigure}
    \vskip\baselineskip
    \begin{subfigure}[]{\textwidth}
        \includegraphics[width=\textwidth]{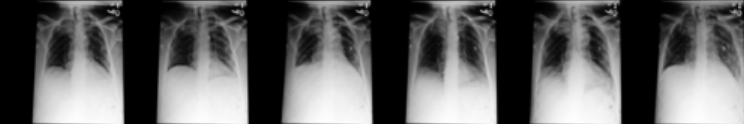}
        \caption{(1) original: black, healthy, 68-year-old female (2) do(age=35) (3) do(diseased) (4) do(white) (5) do(male) (6) do(all).}
    \end{subfigure}
    \caption[]{MIMIC-CXR Counterfactuals from model trained on 20\% labels.}
\end{figure}

\end{document}